# Multiple Abnormality Detection for Automatic Medical Image Diagnosis Using Bifurcated Convolutional Neural Network

M. Hajabdollahi, R. Esfandiarpoor, E. Sabeti, N. Karimi, K. Najarian,,
S.M.R. Soroushmehr, S. Samavi

*Abstract*— **Automating classification and segmentation process of abnormal regions in different body organs has a crucial role in most of medical imaging applications such as funduscopy, endoscopy, and dermoscopy. Detecting multiple abnormalities in each type of images is necessary for better and more accurate diagnosis procedure and medical decisions. In recent years portable medical imaging devices such as capsule endoscopy and digital dermatoscope have been introduced and made the diagnosis procedure easier and more efficient. However, these portable devices have constrained power resources and limited computational capability. To address this problem, we propose a bifurcated structure for convolutional neural networks performing both classification and segmentation of multiple abnormalities simultaneously. The proposed network requires less hardware compared to having separate structures for classification and segmentation. It is trained first by each abnormality separately and then by using all abnormalities. In order to reduce the computational complexity, the network is redesigned to share some features which are common among all abnormalities. Later, these shared features are used in different branches to segment and classify the abnormal region of the image. Finally, results of the classification and segmentation are fused to obtain the classified segmentation map. The proposed framework is simulated using four frequent gastrointestinal abnormalities as well as three dermoscopic lesions and for evaluation of the proposed framework the results are compared with the corresponding ground truth map. Properties of the bifurcated network such as low complexity and resource sharing make it suitable to be implemented as a part of portable medical imaging devices.**

## I. INTRODUCTION

Automatic segmentation and classification of abnormalities in different body organs have important role in medical imaging applications. In different medical imaging applications such as endoscopic, dermoscopic, and fundoscopic imaging, automatic segemtnation and classification of abnormalities greatly help physicians in the diagnosis process. Automatic medical image processing also reduces the time spent by medical experts for analysis of images and video frames. Automatic detection of abnormalities in WCE images has been investigated as multiple or single abnormality detection problems [1]. Different abnormalities in *wireless capsule endoscopy* (WCE), funduscopic images, and different lesions in dermoscopic images are examples of multiple abnormalities which are extensively studied to be automatically detected. In endoscopic imaging, WCE, a non-invasive modality for remotely screening of all parts of human's *gastrointestinal* (GI) system, can be used in detecting different types of abnormalities such as bleeding and Crohn's disease [2] [3]. In skin lesion analysis, dermoscopy can be used to discriminate between benign and malignant skin lesions and to improve the diagnostic capability [4].

Jia and Meng proposed a single abnormality detection method in which a *convolutional neural network* (CNN) architecture is employed for classification of bleeding images [1]. In [5], features extracted by CNN as well as features from K-means clustering are used for classification of bleeding images. Vieira *et al.* [6], proposed an accelerated expectation maximization method called *maximum posterior* to classify images into angiodysplasia and other abnormalities. Moreover, they used Markov random field to take spatial information into account. Fu *et al.* [7], proposed a method for bleeding regions detection in WCE images. In their work, after removing detected edge pixels, superpixels are created by a grouping method. Finally, superpixels are classified using a *support vector machine* (SVM) approach. In [8], initial seeds detected using K-means are labeled by SVM, then bleeding regions are segmented using an interactive segmentation based on cellular automata. In [9], WCE images are represented using local binary pattern and

Mohsen Hajabdollahi, Reza Esfandiarpoor, and Nader Karimi are with the Department of Electrical and Computer Engineering, Isfahan University of Technology, Isfahan 84156-83111, Iran.

E. Sabeti is with the Department of Computational Medicine and Bioinformatics, University of Michigan, Ann Arbor, U.S.A.

S.M. Reza Soroushmehr is with the Michigan Center for Integrative Research in Critical Care, and also with the Department of Computational Medicine and Bioinformatics, University of Michigan, Ann Arbor, U.S.A.

Kayvan Najarian is with the Department of Computational Medicine and Bioinformatics; Department of Emergency Medicine; and the Michigan Center for Integrative Research in Critical Care, University of Michigan, Ann Arbor, U.S.A.

Shadrokh Samavi is with the Department of Electrical and Computer Engineering, Isfahan University of Technology, Isfahan 84156-83111, Iran. He is also with the Department of Emergency Medicine, University of Michigan, Ann Arbor, U.S.A.

 

average saturation and are classified using SVM in bleeding and non-bleeding regions. In [10], color information of selected points by *speed up robust features* (SURF) are extracted and classified as lesion and non-lesion regions. In this study different types of abnormalities are considered as one class and healthy parts as the other one.

Also in dermoscopic images, in [11], a pigment network is extracted based on 2-D Gabor filtering and also more features are extracted based on color, texture and geometric properties. At last, neural network and SVM methods are applied to classify images as benign and malignant. Ruela *et al.* [12], segmented dermoscopic images and extracted a subset of the best shape and symmetry features using a filter approach method. Finally, images classified as melanoma and non-melanoma using KNN classifier. Nasir *et al.* [13], after hair removal and contrast enhancement, extracted abnormal regions of the image using active contour and uniform distribution based segmentation. In their work, in the segmented area, color and texture features are extracted and classified using SVM classifier. In [14], main frequencies of FT are used as input features for classification of melanoma. Moreover, a co-occurrence matrix is created using main frequencies of FT and then images are classified by applying a *multi-layer perceptron* (MLP) and SVM. In [15], skin lesions are segmented using fully convolutional neural network by defining a loss function that takes imbalance data problem into account. In [16], in the first stage of the proposed multi-stage fully convolutional neural network, a fully convolutional neural network conducts the segmentation task. In next stages, segmentation is performed using estimated segmentation map resulted from the previous stage. In [17], skin lesions are segmented to different regions based on their color and texture using a graph based method. Each segmented region is labeled using a method named corr-LDA. After that lesions are classified using various classifiers including random forest, KNN, and SVM. In [18], a skin lesion segmentation method is proposed where hairs in skin images are removed and a region of interest is segmented using a region growing approach. In [19], after segmentation of lesions, 2-D and 3-D representation of images are reconstructed. Then, classification is performed using different features including shapes of 2-D and 3-D reconstruction as well as texture and color. In [20], after using an active contour method for segmentation, lesions are labeled by SVM. In [21], skin areas are detected using a thresholding method and lesion regions are segmented using a Delaunay triangulation method. Final segmentation map is obtained by fusing results of lesion segmentation and skin detection step.

Since there are different types of abnormalities in GI and other parts of human body, the problem of detecting multiple abnormalities has been addressed in a number of studies. Seguí *et al.* [22], designed a CNN architecture for detecting different GI events such as bubbles, wrinkles, and others. Yuan *et al.* [23], utilized an SVM approach for classifying polyps, ulcer, and other abnormalities using three main features extracted from original images. These features include color and texture features clustered by K-means, saliency map obtained through color and texture, and local manifold structure resulted from nearest neighbor graph. A single CNN structure is used by [24] for classification of body organs including stomach, small intestine, and colon. Iakovidis *et al.* [25], detected bleeding regions by defining a number of criteria on super pixel saliency. In [26], a robust modified speeded-up feature is used to include color features which are important in WCE images. Then SVM is used for classification of abnormal and normal images. A single CNN structure is used as a binary classifier for detection of different abnormalities separately by Sekuboyina *et al.* [27]. Deeba *et al.* [28], detected salient regions using conditional probability of a region around each pixel and also identified different types of abnormalities using adaptive thresholding. Vasilakakis *et al.* [29], took advantage of clustering algorithms and SURF to extract a bag of visual words. Finally, a trained SVM was applied to detect inflammatory lesions. In [30], semantic information of WCE images is used for better lesion classification. The difference of maximum points is used as a bag of words for salient point detection, and in order to detect different objects in WCE images, multi-label classification is used.

Developing a battery operated handheld device with capability of automatic segmentation and classification of medical abnormalities is very demanding [31]. Development of methods for skin cancer screening using devices such as smartphones and portable cameras for skin screening, has provided efficient diagnosis services at a very low cost. Simple and efficient diagnosis procedures should be designed for portable devices which are small and have limited battery life-time. In [32] and [33], a lightweight method for melanoma detection in smartphones is developed. Also in [34], a device for automatic real-time screening of skin is designed to classify region of interest as benign and malignant. In [35], a handheld system for automated detection of melanoma is designed. In [36], a hardware system for analysis of skin lesions is proposed and implemented on an embedded platform.

In the endoscopy process conducted by a capsule, images should be transmitted outside the capsule. Hence, summarization and compression are introduced as proper ways to reduce the amount of transmitted data [37]–[39]. In [40], bleeding regions are detected using an *artificial neural network* (ANN) structure which is simplified to become suitable for hardware implementation inside the capsule. In [41], an FPGA implementation of data processing core with a focus on image compression for WCE device is presented. In [42], a low complexity architecture for one-dimensional *discrete cosine transform* (DCT) transform is presented. In [43], concerning low-resolution and properties of WCE images, specific patterns in different image channels are extracted for DPCM coding and are utilized for better compression.

In the scope of implementing automatic medical image processing methods inside portable devices, there are two main concerns. First, a few research studies have investigated the problem of detecting multiple abnormalities inside the portable medical devices such as digital dermatoscope and WCE device. Second, there is no simple and efficient CNN structure for simultaneous segmentation and classification of multiple abnormalities. Taking these considerations into account can be very beneficial for better and easier diagnosis in medical portable device.

To address the aforementioned problems, in this paper, a new algorithm for segmentation and classification of various abnormalities in different medical applications is presented.

     3

The proposed bifurcated neural network segments images in one branch and classifies them in the other branch. Thanks to existence of common low-level features in all abnormalities, the main part of the proposed network is designed to be the same in both branches and share computational operations. The contributions of the proposed method can be regarded as follows:

- Design a simple network structure for both segmentation and classification of different abnormalities in medical images.

- Design a bifurcated network with a shared primary part that extracts common features which are useful for both segmentation and classification of different abnormalities.

- Reuse resources in the primary part of the proposed network that leads to a simple network with low structural complexity suitable for portable medical devices.

The remainder of this paper is organized as follows. In Section II, different structures for classification of multiple abnormalities are presented. In Section III, the proposed bifurcated network for detection of multiple abnormalities is presented. Section IV is dedicated to experimental results, and finally, in Section V the concluding remarks are presented.

## II. MULTIPLE ABNORMALITY DETECTION

Classification and segmentation of multiple abnormalities are used in many medical applications such as endoscopic, dermoscopic, funduscopic, etc. Abnormalities such as microaneurysms and exudate in retinal images and bleeding, ulcer and chylous in GI images are examples of multiple abnormalities related to important diseases. In recent years CNN has become very popular for efficient automatic detection of abnormalities in medical imaging applications. Multiple abnormalities can be detected using different CNN structures which are briefly explained as follows.

### A. Single network structure for multiple abnormality detection

A straightforward solution for multiple abnormality detection is utilizing a single structure for all types of abnormalities. In Fig. 1, a sample patch-based CNN structure for abnormality detection in endoscopic images is illustrated. In this structure, the network is trained using all types of abnormalities and classifies an input pixel into one of the abnormality classes according to its surrounding patch. In Fig. 1, classification is performed based on the surrounding patch of each pixel, and this classification can be utilized for classification and segmentation of the whole image. Single structure of neural network has been used in recent years to detect body organs and abnormalities. As mentioned earlier, in [10, 12, 15], the problem of classification of WCE images was addressed. Also classification of multiple abnormalities is investigated in other areas such as retinal image analysis [26-27]. In [44], a single CNN structure is used for segmentation of optic disk, fovea, and blood vessels. In [45], a single CNN structure is utilized for segmentation of exudates, bleeding, and micro-aneurysms. Although a single structure is utilized in many research areas, the resulting network tends to be computationally complex. For example in [1] and [22], about $15 \times 10^6$ and $2.2 \times 10^6$ parameters are utilized, respectively. Having so many parameters is troublesome for implementation of an algorithm inside a portable medical device. Also during training, the single network structure, each abnormality modifies the whole network variables which can cause poor classification performance specially when abnormalities have a lot of variations.

Designing a simple and adequately accurate CNN structure for detecting multiple abnormalities inside portable devices is a challenging task. Moreover, in some situations, abnormalities are very different in terms of color, size, and feature. Hence, it is not possible to utilize an efficient single structure for detecting all types of abnormalities. Assigning special parts of the network to each abnormality can be regarded as an efficient way for better training the network. In this manner, each part of the network is trained and customized for a special abnormality and could be exclusively improved and optimized.

### B. Separate network structure dedicated to each abnormality

In Fig. 2, separate networks are dedicated to each abnormality where each network is trained using only one abnormality image dataset. Each network can accurately segment the abnormality used during the training phase. In this manner, each network structure is better specialized and customized on an individual abnormality, but it is not able to indicate the abnormality class.

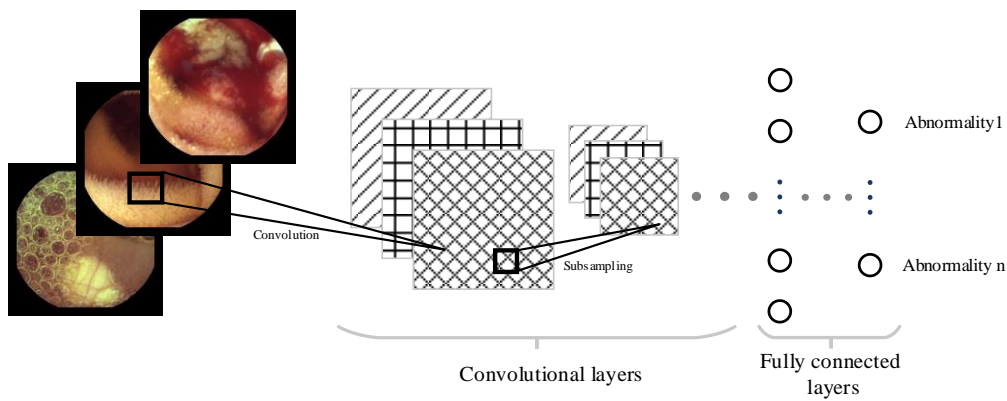

Fig. 1. Single CNN structure for multiple abnormality detection

 

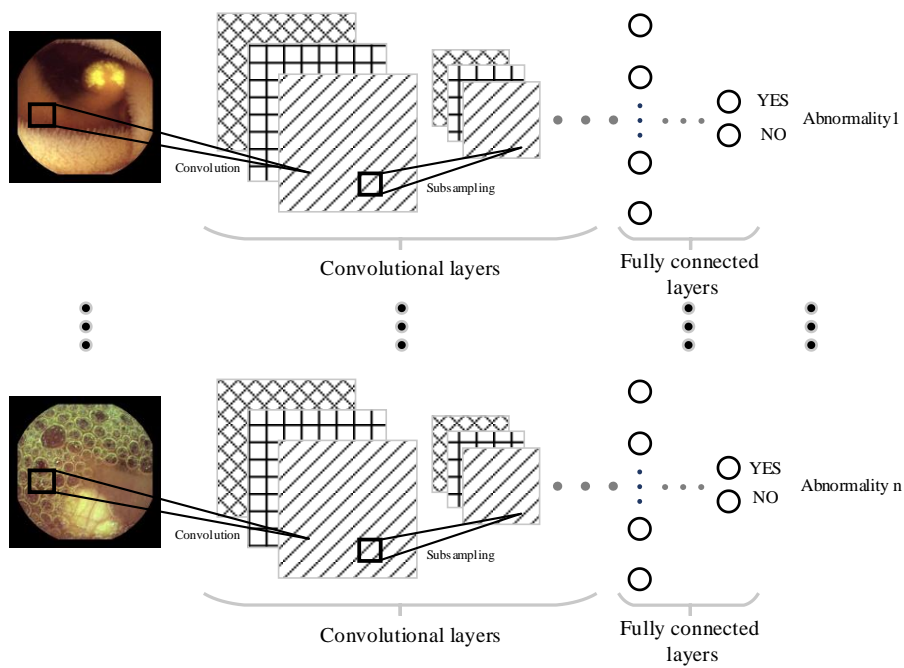

Fig. 2. Separate networks dedicated to each abnormality.

In these networks, there is no cooperation among elements of separate networks and similar features among abnormalities are not utilized which increases the overall complexity of the network structure. Although these networks are specialized adequately, they suffer from redundant operations. Furthermore, they are only useful for segmentation and do not help in the classification task. Concerning multiple abnormalities segmentation and classification, there must be a possibility to concentrate on each abnormality as needed for segmentation without loss of generality required for classification.

## III. Proposed Bifurcated Network for abnormality classification and segmentation

In Fig. 3, two different abnormalities in WCE and dermoscopic images are depicted with their corresponding ground truth, respectively. There are various abnormalities that are common in human GI and skin which can be related to important diseases. For better diagnosis, both segmentation and classification of abnormal regions are required. Single network structure is not specialized for individual abnormalities. Hence, it is not always possible to appropriately train the network to reach a proper performance. Moreover, separate networks for each abnormality are not able to classify the images into abnormality classes and suffer from redundant operations. In the following subsections, the proposed network addresses these challenges.

### A. Overview of the proposed framework

In Fig. 4, a general block diagram of the proposed framework is shown. We first train separate neural networks where each one is dedicated to a specific abnormality and can segment it precisely. Then, the first parts of all the trained segmentation networks are merged to make the primary part of the final network. The principal idea here is that when the primary parts of the several neural networks are merged, the secondary parts are needed to be retrained exclusively to retain the same functionality as before. Let's talk about this claim in more detail for a simple example of $n$ different 3-layer MLP networks. A neural network in general can be interpreted as a type of Markov random field, with a layer of neurons as a vector of stochastic visible (input) unit $\boldsymbol{v}$ and vectors of various stochastic hidden units $\boldsymbol{h}$. In our example of Fig. 5, suppose each network is trained on a specific class of inputs, resulting in $n$ different sets of weight matrices, $\mathcal{W}^1, \mathcal{W}^2, \cdots, \mathcal{W}^n$, where $\mathcal{W}^i = \{W_1^i, W_2^i, W_3^i\}$ is the set of weight matrices of 3 layers trained on $i$th input class (abnormality). In this setup, the outputs of networks are

$$\begin{cases} \boldsymbol{y}^1 = \boldsymbol{h}_3^1 = W_3^1 W_2^1 W_1^1 \boldsymbol{v}^1 \\ \boldsymbol{y}^2 = \boldsymbol{h}_3^2 = W_3^2 W_2^2 W_1^2 \boldsymbol{v}^2 \\ \quad\quad\vdots \\ \boldsymbol{y}^n = \boldsymbol{h}_3^n = W_3^n W_2^n W_1^n \boldsymbol{v}^3 \end{cases} \quad (1)$$

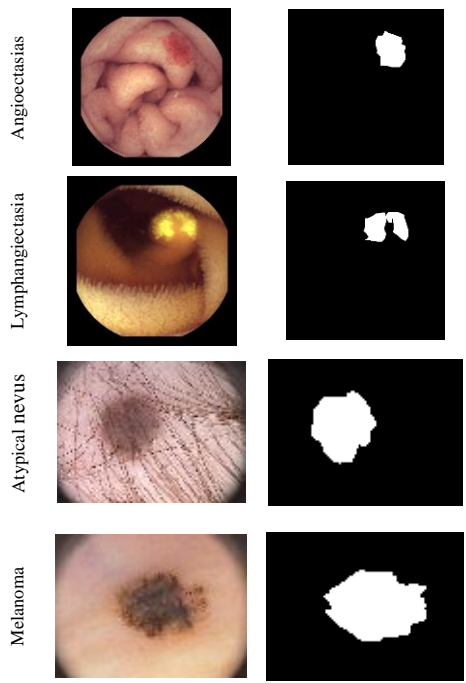

Fig. 3. Four examples of abnormalities in endoscopic and dermoscopic images and corresponding Ground Truth

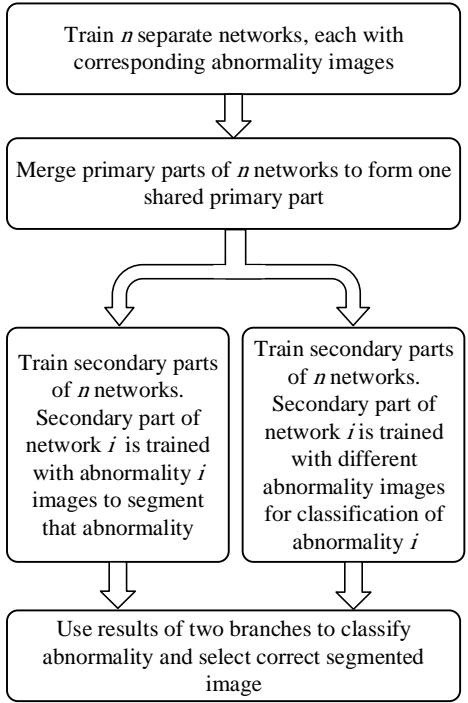

Fig. 4. Main steps of the proposed network formation framework.

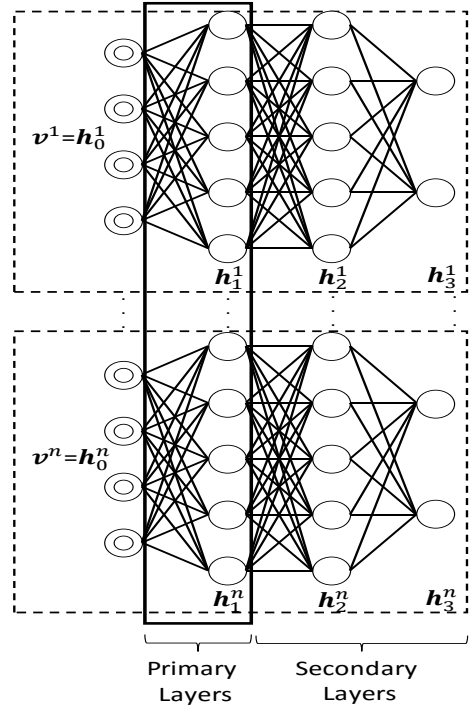

Fig. 5. Neural networks trained on specific class of inputs

As mentioned before, since low-level features in all abnormalities are common, the parameters of first layer (primary) of the trained neural networks can be merged (i.e. the parts inside the box of Fig. 5, with solid line) to eliminate redundant computations and save memory. As such, suppose we merge the weights of the first layer (in the box with solid line) of all the $n$ trained networks. In this case, the network of Fig. 6, can be an alternative with significant reduction of memory requirement and computations, and the goal is to design it in such a way that the output of both networks could be the same. For this network, we can assume that the weight matrix of the common primary layer is $W_1 = W_1^j + \Delta W_1^j$ where $1 \le j \le n$. Thus the $j$th output for input $\boldsymbol{v}^j$ would be

$$\begin{aligned} \widehat{\boldsymbol{y}^j} &= W_3^j W_2^j W_1 \boldsymbol{v}^j \\ &= W_3^j W_2^j (W_1^j + \Delta W_1^j) \boldsymbol{v}^j \\ &= W_3^j W_2^j W_1^j \boldsymbol{v}^j + W_3^j W_2^j \Delta W_1^j \boldsymbol{v}^j \\ &= \boldsymbol{y}^j + W_3^j W_2^j \Delta W_1^j \boldsymbol{v}^j \end{aligned} \quad (2)$$

Therefore, in order to have the same output for both networks, i.e. $\widehat{\boldsymbol{y}^j} = \boldsymbol{y}^j$ (in element-wise sense), the secondary layers should be retrained.

Now going back to Fig. 4, the primary part of the new network is then followed by two secondary branches of the final

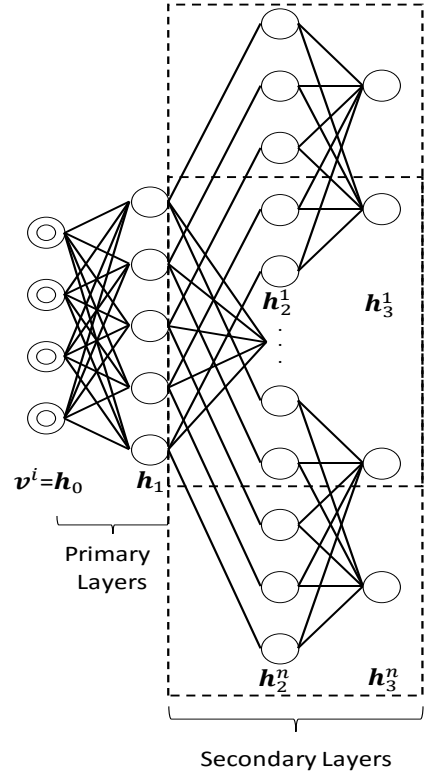

Fig. 6. Merging primary layers of separate neural networks

 6

network. One of these secondary branches performs the classification task, and the other one is dedicated to the segmentation task. In the segmentation branch of the network, each secondary sub-network is exclusively trained using a specific abnormality to precisely segment that abnormality. Also, in the classification branch of the final network, each secondary sub-network is trained to identify a specific abnormality. Finally, segmentation results are fused with classification results and will make the final classified segmented map. Different parts of Fig. 4, are explained in more details in the following.

### B. Separate network structure for each abnormality

As it was mentioned in section II-B, a straightforward solution for segmentation can be realized by assigning separate networks to each abnormality. In this way, separate networks are trained on images containing only one abnormality, and each network is responsible for segmentation of the corresponding abnormality's images. In order to evaluate performance of separate network structures for each abnormality, an experiment is conducted using MNIST dataset. MNIST is an image dataset containing 10 classes of English handwritten numbers. 50,000 and 10,000 image samples are used for the network training and testing respectively. As illustrated in Fig. 7a, a separate network is dedicated to one class of numbers. Each network is a CNN structure with 64 and 32 filters in each *convolutional layers* (CLs) and with 200 and 100 neurons in each *fully connected layers* (FCLs). Each network is responsible to discriminate between its own class and other classes of numbers. In this structure, each network is trained to classify images of its own class as "POSITIVE" and other nine classes as "NEGATIVE". Results of each network are shown in Table I. Although for each class of numbers proper results are obtained, this structure cannot be used for classification of all classes of numbers. To have a network with classification capability, an input image with unknown number's class should be applied on all ten networks and after that a procedure must be considered to vote among results of different networks. In addition to the classification problem, ten separate networks for ten numbers lead to a network with extreme structural complexity. Generally, for medical applications, this approach could lead to complex structures and redundant operations. Besides, we need to know the type of abnormality before feeding the medical image into an appropriate network for segmentation.

Table. I. Result of separate network structures dedicated to each class of numbers on MNIST dataset

| Class of numbers | Accuracy |
|---|---|
| 0 | 99.94 |
| 1 | 99.90 |
| 2 | 99.86 |
| 3 | 99.90 |
| 4 | 99.89 |
| 5 | 99.87 |
| 6 | 99.82 |
| 7 | 99.77 |
| 8 | 99.89 |
| 9 | 99:69 |

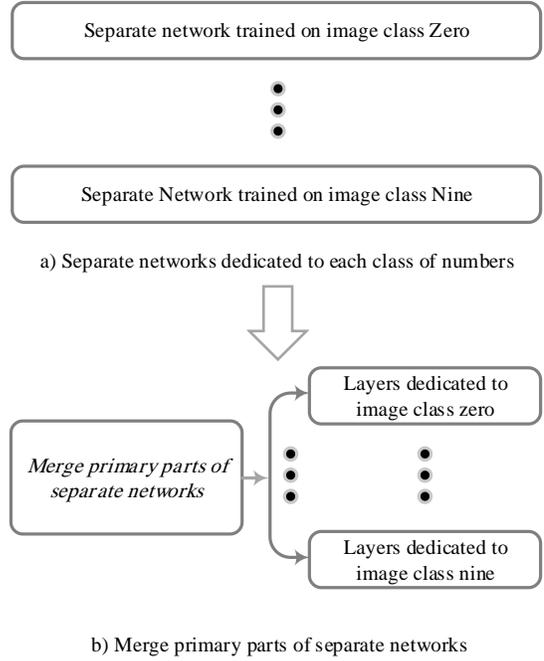

a) Separate networks dedicated to each class of numbers

b) Merge primary parts of separate networks

Fig. 7. Training separate network and merging primary parts of separated network on MNIST dataset.

### C. Merge primary parts of separate networks

There are similarities among all images from multiple abnormalities at least in the normal parts of these images. General features extracted from all images could be assigned to a shared part of the final network. Using a network with a shared part, redundant computations can be eliminated and network structure would be simplified. As illustrated in Fig. 7b, separate networks for each class of numbers in MNIST dataset are merged with each other to share primary parts of their structures with each other. To discriminate between images from different number's classes, a secondary part is considered for each class of numbers named as class-specific (abnormality-specific) layer to detect number's class. For MNIST dataset, the primary part consists of two CLs with 64 and 32 filters and secondary parts for each number consists of FCLs with 200 and 100 neurons. Results of the network with the shared primary parts (i.e. merged network) are illustrated in Table II.

In order to validate our method, we compare its results with a monolithic structure for MNIST dataset commonly used for comparing MNIST classification algorithms. In the monolithic network a single structure is dedicated for all 10 number's classes and results are illustrated in Table II. Monolithic network is a CNN structure with 64 and 32 filters in CLs and with 200 and 100 neurons in FCLs. As illustrated in Table II, classification accuracy of 99.46% and 99.23% are achieved for monolithic and merged networks respectively. It can be observed that for the MNIST classification problem, primary parts of the networks for each class of numbers can be merged. Also, in the secondary part, separate networks can be considered for each class of numbers. Although for the classification of MNIST dataset, merging operation does not lead to increase in performance, but grants classification capability to the separated networks for each number and reduces total network complexity.

Table. II. Classification result of different network structures on MNIST image dataset

| Network structures | Accuracy |
|---|---|
| Monolithic | 99.46 |
| Merged | 99.23 |

In medical applications where both classification and segmentation are required, merging operations reduces the computational complexity. When we look at the merged network for multiple abnormality detection in WCE, the primary parts of the networks, dedicated to the extraction of general features, are similar among different networks. The secondary part of the network is dedicated to each specific abnormality. In the proposed merged network, variables in the primary part are trained on all abnormalities. Moreover, the networks in the secondary parts are trained on each abnormality separately. Networks' layers in the primary part can be merged in different ways depending on the types of abnormalities. In a simple schema, the primary layers of each separate network are designed to have the same structure, and their weights can be averaged to make the primary part of the final merged network. In Fig. 8, the proposed bifurcated structure for multiple abnormality detection is illustrated where the shared primary part is bifurcated into two main branches. One branch is dedicated to abnormality-specific segmentation, and the other one performs abnormality classification. The primary part of the merged network is used once during the training of the segmentation path of the network and then during the training of the classification branch of the bifurcated network. In the next two sub-sections, the primary part of the network is used for segmentation and classification respectively.

### D. Training the segmentation branch

In the experiment conducted on MNIST, it was observed that primary parts of the separated networks were merged and 10 networks were dedicated to each class of numbers. This experiment shows that there are primary parts in MNIST which are common in all images. Also, in order to specify different classes of numbers, there is an essential need to consider a secondary part for each class of numbers. In this section we use merging operation as we conducted in MNIST experiment to make bifurcated network for segmentation and classification. As illustrated in Fig. 8, the secondary part of the final merged network in segmentation branch consists of multiple sub-networks, called abnormality-specific layers, which are dedicated to each abnormality exclusively. The secondary parts of separate networks could make these sub-networks. Primary part of the network is trained on all image abnormalities. Since the primary parts of the separated networks are now merged, the abnormality-specific layers should be re-trained. The abnormality-specific layers are exclusively trained for each abnormality. In this manner, in the final merged network, the shared low-level features are produced in the primary part and high-level specialized features for each abnormality are generated by the abnormality-specific layers located in the secondary part of the network. When an image with abnormality of type "$i$" is sent to all $n$ segmentation netwroks, only network $i$ is expected to perform accurate segmentation and other segmentation networks have invalid segmentation results for that image. All networks of the segmentation branch produce segmentation maps for abnormality $i$. The classifying branch will select the output of network $i$.

### E. Training the classification branch of the network

Generally, forming a single network for classification of all abnormalities is either impossible or it can only be achieved by a very complex structure. Hence, we reused the primary part of the final merged network to train another set of abnormality-specific layers for classification as illustrated in Fig. 8. We use all types of abnormality images to train each of these $n$ networks. When the network $i$ receives other abnormality images it considers them as normal. Network $i$ is trained to

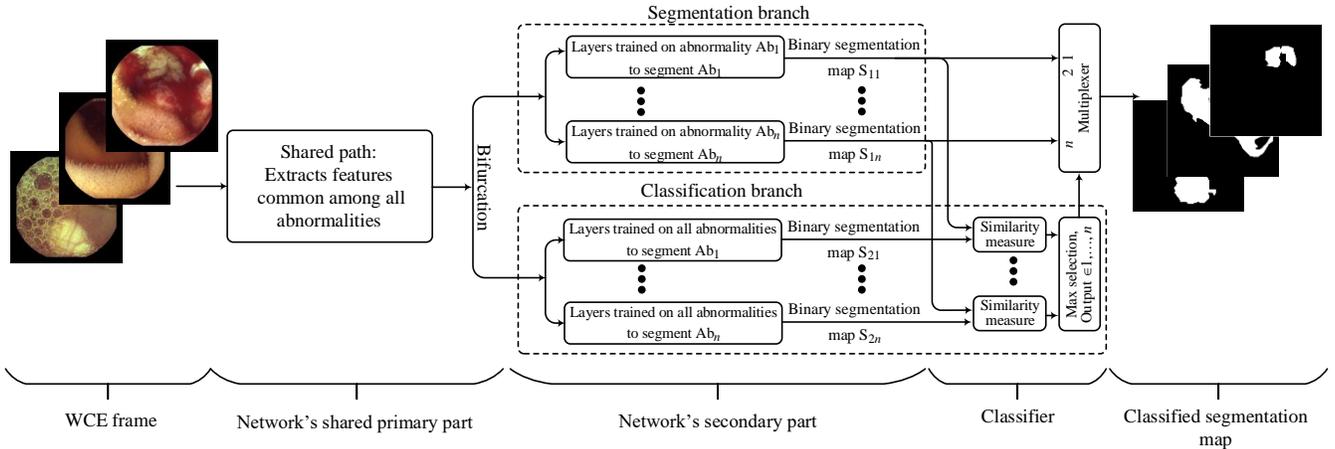

Fig. 8. Proposed bifurcated network structure for multiple-abnormality detection.

 

segment only its own abnormality type. Hence, this type of segmentation contains implied classification.

As illustrated in Fig. 8, the secondary parts of all networks in the classification branch are trained with different abnormalities. The primary part of the network is the same shared primary part of the merged segmentation network. Each abnormality-specific layer in classification branch is dedicated to a specific abnormality but is trained using all abnormalities. Segmentation branch produces a set of binary segmentation maps defined as $S = \{S_1, ..., S_n\}$. Also the networks in the classification branch produce a set of binary classification maps defined as $C = \{C_1, ..., C_n\}$. We can select the best $S_i$ if we know which type of abnormality is being segmented. We propose that all pairs of binary segmentation maps, $\{S_i, C_i\}_{i \in 1, ..., n}$, should be compared by calculating their similarities using (1) where "|.|" refers to cardinality operator.

$$similarity(i) = \left( \frac{|S_i \cap C_i|}{|S_i|} + \frac{|S_i \cap C_i|}{|C_i|} \right) / 2 \qquad (3)$$

Those pixels detected as abnormality by either the segmentation branch or the classification branch are represented as 1's in the binary intersection map. The $k$th pair is selected such that

$$k = \arg\max(similarity(j))$$
$$j \in \{1, ..., n\} \qquad (4)$$

The intersection of the two binary maps of $S_i$ and $C_i$ forms a binary "intersection map". The selection of the $k$th pair of intersection map that has the highest similarity, based on (4), indicates that the image has abnormality $k$ and segmentation map $S_k$ will be selected as the output of the network. When the left operand in the addition operation of (3) is close to 1 it indicates that the two maps are similar to each other and are similar to the map produced in the segmentation branch of the network. When the right addition operation of (3) is close to 1 it shows that the maps are similar with each other and are similar to the map generated by the classification branch of the network. The ratio for the number of ones in the intersection map to the number of ones in the segmentation map demonstrates the correspondence between segmentation and classification with respect to the segmentation. This similarity level is expressed by the left operand of the addition operation of (3). Also, the right operand in the addition operation of (3) demonstrates the correspondence between the segmentation and classification with respect to the classification. The similarity metric defined in equation (3), demonstrates the analogy between the classification and segmentation results. The abnormality which maximizes the similarity between corresponding segmentation and classification results is selected as the abnormality class. After determining the abnormality type, corresponding segmentation is considered as the final segmentation result.

## IV. EXPERIMENTAL RESULTS

The proposed network structure for classification and segmentation of multiple abnormalities in medical images is implemented using TensorFlow framework. A PC equipped with an Intel(R) Core(TM) i7-4790 CPU 4.00 GHz and 32GB of RAM and NVIDIA GeForce GTX TITAN X is used for the training phase and testing the network performance. The bifurcation approach reduces the number of operations and simplifies the network structure which makes it suitable for applications that require simple operations. Here, two applications including WCE abnormality and dermoscopic skin lesion detection are experimented. Multiple abnormality segmentation and classification in WCE and dermoscopic images are very useful for better diagnosis and reducing time spent by physician for diagnosis. Capsule endoscopy is a tiny device with limited hardware resources and wireless dermatoscope is a battery operated portable device. Hence, implementing a simple network structure and reusing operations could reduce required resources. The bifurcation approach is implemented as one network structure which is trained and tested on the WCE abnormality images. KID and bleeding data sets are considered and images including bleeding, angioectasias, chylous and lymphangiectasia are used from [40],[46] and. Images used for training and testing of WCE include 50 bleeding, 27 angioectasias, 8 chylous and 9 lymphangiectasias images. For better evaluation, the bifurcation approach is implemented as another network structure which is trained and tested on the skin lesion image data set (PH2) including common nevus, atypical nevus and melanoma which is publicly available at [48]. PH2 includes 80 common nevus, 80 atypical nevus and 40 melanoma images.

We used accuracy, sensitivity and specificity for evaluation of classification and segmentation. In order to better evaluate segmentation results, we used Dice score which is an appropriate metric for datasets with imbalance data problem. Overlapping image patches and their central pixels are used as network's inputs and desired label, respectively during the training process. WCE input images are converted to three different color channels including gray-scale level, "S" channel of HSV color space, and '$a$' channel of CIE-lab color space and RGB channels are used for PH2 images. For each of the three different channels, patches with the size of 9×9 are selected as network inputs, and the central pixel is labeled according to the corresponding ground truth. Training data is balanced in such a way that the number of healthy patches is twice the number of abnormal patches. Simulation results are organized in the following subsections.

### A. Single network structure for classification of all abnormalities

Two single structures of the convolutional neural network are applied for classification of abnormalities in WCE and dermoscopic images. We tried different single structures with different convolutional and fully connected layers, and selected the best one which has two CLs with 64 and 32 kernels as well as two FCLs with 60 and 40 neurons. Classification is performed for each pixel based on a patch-wise approach, and visual results of six WCE and dermoscopic images containing different abnormalities are illustrated in Fig. 9 and Fig. 10, respectively. In Fig. 9, classification results of bleeding, angioectasias, chylous and lymphangiectasia are colored with red, yellow, green and blue respectively. Also in Fig. 10, classification results of common nevus, atypical nevus and melanoma are colored with green, blue and red respectively. Visual results of classifications show that single network structure is not able to properly classify the abnormalities. In images in Fig. 9 and Fig. 10, some regions are wrongly classified. Although the single network is not able to distinguish different abnormality classes, the abnormal parts of the corresponding abnormality are segmented approximately. For example, in images with angioectasias abnormality in Fig. 9, angioectasias parts are almost colored yellow correctly. Also, in common nevus images of Fig. 10, common nevus lesion parts are almost colored green correctly. Although there are other colored regions in the result of segmentation, regions with the correct abnormality are similar to the ground truth approximately.

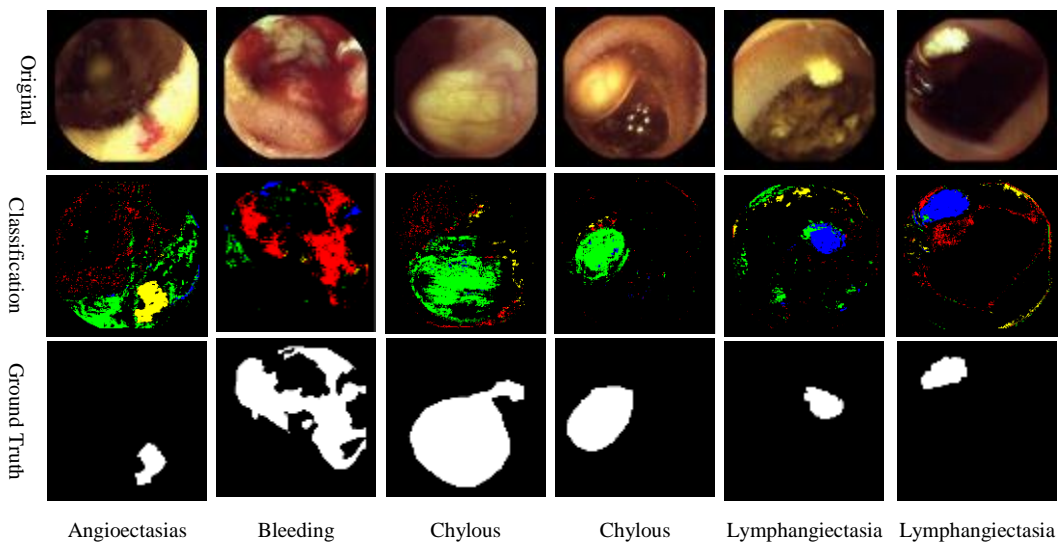

Fig. 9. Visual results of the single classification network for WCE images

### B. Separate network structures for each abnormality

Here we set up separate network structures for segmentation of each abnormality. Since the primary parts of these networks are going to be merged, these parts are selected to have the same configuration in primary part.

For WCE and dermoscopic images, four and three networks are trained respectively with a set of images containing only one abnormality. These networks have two CLs with 64 and 32 convolutional filters and two FCLs with hidden layers containing 60 and 40 neurons. For evaluation of each separate network, segmentation is performed only on images of corresponding abnormality. Visual results on sample WCE and skin lesion images are illustrated in Fig. 11 and Fig 12 respectively. Segmentation results represent that proper segmentation is obtained using separate networks dedicated to each abnormality. Dice score and accuracy of each segmented abnormality are provided in Table III and Table IV. Since each network is trained on a specific abnormality, it is not able to discriminate among different abnormality classes. Although acceptable segmentation results are obtained, this structure is not able to classify the images into proper classes.

### C. Separate networks with merged primary parts

As merging primary layers of separate networks reduces computational operations and simplifies the network's structure, we merged the convolutional layers of all separate networks to make a new segmentation network with the merged primary layer. Then, we trained primary part of merged network as common layers and trained secondary part as abnormality-specific layers for each abnormality. For WCE and skin lesion images four and three abnormality-specific layers are considered respectively. Abnormality-specific layers in WCE and skin lesion networks have two FCLs with 60 and 40 neurons. Quantitative results of the segmentation network with merged primary part for WCE and skin lesion image dataset are illustrated in Table V and Table VI respectively. Table V and VI, indicate that there is no significant difference from the previous separate networks' results in Table III and Table IV respectively. As it was observed in MNIST experiment, merging primary parts of separate networks reduces network complexity with no significant loss of network performance.

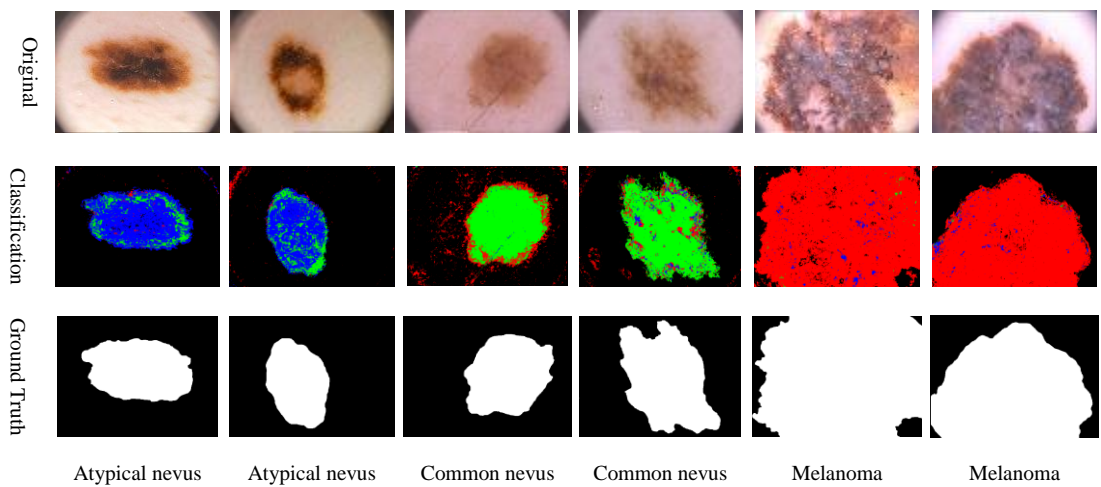

Fig. 10. Visual results of the single classification network for dermoscopic images

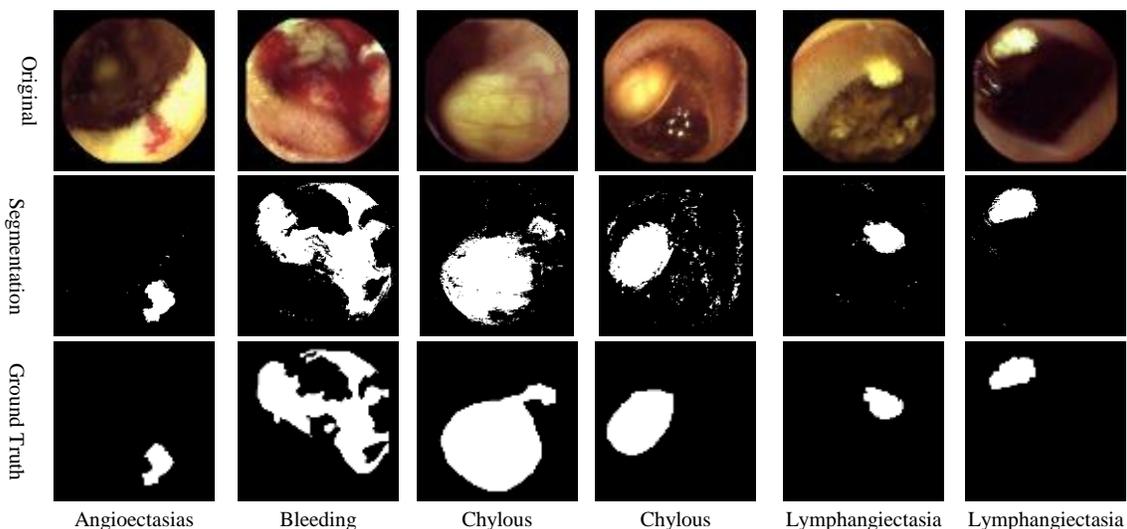

Fig. 11. Visual results of the segmentation with separate networks for WCE images

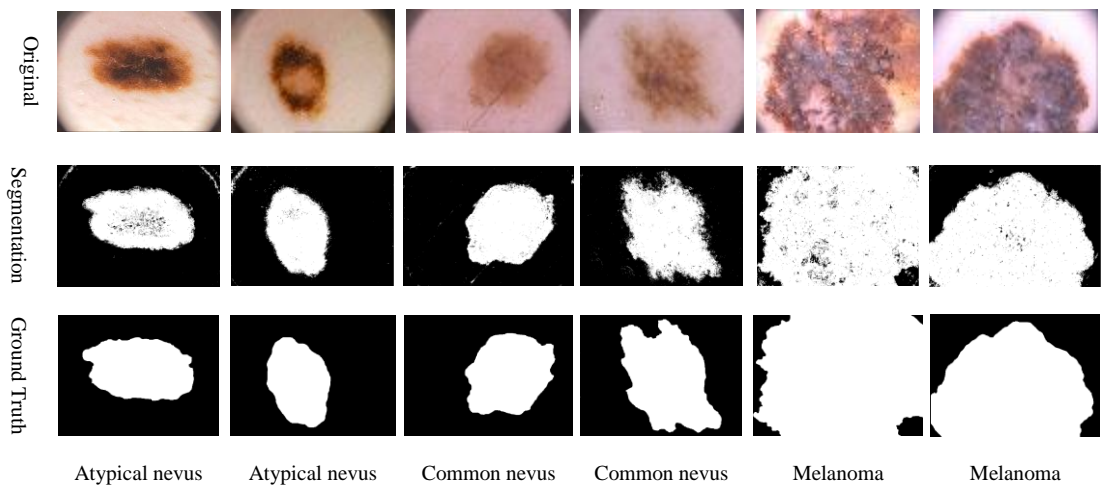

Fig. 12. Visual results of the segmentation with separate networks for dermoscopic images

Table III. Quantitative results of segmentation using dedicated network for each WCE abnormality

| Abnormality | Accuracy | DICE |
|---|---|---|
| Angioectasias | 0.99 | 0.76 |
| Bleeding | 0.94 | 0.86 |
| Chylous | 0.96 | 0.86 |
| Lymphangiectasia | 0.99 | 0.94 |

Table IV. Quantitative results of segmentation using dedicated network for each skin image abnormality

| Abnormality | Accuracy | DICE |
|---|---|---|
| Common nevus | 0.973 | 0.940 |
| Atypical nevus | 0.960 | 0.930 |
| Melanoma | 0.951 | 0.962 |

Table V. Quantitative results of segmentation using merged network for all WCE image abnormalities

| Abnormality | Accuracy | DICE |
|---|---|---|
| Angioectasias | 0.99 | 0.76 |
| Bleeding | 0.91 | 0.85 |
| Chylous | 0.95 | 0.85 |
| Lymphangiectasia | 0.99 | 0.93 |

Table VI. Quantitative results of segmentation using merged network for all skin image abnormalities

| Abnormality | Accuracy | DICE |
|---|---|---|
| Common nevus | 0.972 | 0.939 |
| Atypical nevus | 0.956 | 0.922 |
| Melanoma | 0.944 | 0.957 |

In the problem of multiple abnormality detection in medical applications, we employed the primary parts in another branch for classification task.

### D. Bifurcated network for classification and segmentation

The network structure in the previous section had primary and secondary parts used for segmentation. As illustrated in Fig. 8, the primary part of this network is used in another path for the classification, and a secondary part is also considered for classification. For WCE and skin lesion images, four and three abnormality-specific layers are considered respectively. Abnormality-specific layers in WCE and skin lesion networks have two FCLs with 60 and 40 neurons. In Fig. 13 and Fig. 14, classification results for six WCE and six dermoscopic image samples are illustrated respectively. Due to assigning abnormality-specific layers for each abnormality, in Figs. 13 and 14, four and three outputs are respectively indicating the correspondence between image pixels and abnormality/ skin lesion classes.

It can be observed from Fig. 13 and Fig. 14 that classification results in related classes are close to the image ground truth. For example, results of chylous images in part of chylous classes are much similar to the ground truth than the others. In this step the applied network for classification is not similar to the experimented monolithic network. We should indicate that the monolithic network used in section IV-A, which can segment all abnormalities, has a highly complex structure as compared to the proposed bifurcated network. Also, segmentation map resulted from the monolithic network is not as accurate as the result of separate networks for each abnormality. Moreover, the monolithic network is not able to decide between abnormality classes. Also as it was observed in Fig. 10, segmentation results are similarly proper but the network is not able to decide between different abnormality classes. Classified results of the merged bifurcated network are used to obtain the final classified-segmented results. In Fig. 11 and Fig. 13 for WCE images as well as Fig. 12 and Fig. 14, for dermoscopic images, it was observed that results of corresponding segmentation and classification are similar in their own abnormalities. For example, for an angioectasias image in Fig. 13, the classification result in the angioectasias class is similar to the segmentation result of the corresponding image in Fig. 11. Also, melanoma lesion segmentation result in Fig. 12, is similar to the result of melanoma class in Fig. 14. Hence using equation (3), abnormality-specific layer with maximum similarity is selected as the class of the abnormality.

In Fig. 15 and Fig. 16, visual results of the bifurcated network for sample WCE and dermoscopic images are shown respectively. According to our observations, proper classification and segmentation are possible by using the fused results of segmentation and classification branches. Using fused results of the bifurcated network, we can select the best segmentation results from the segmentation branch. Also, the fused results help classify input images properly. In Table VII and VIII, the number of true and false predictions of abnormality classes in WCE and dermoscopic datasets is shown respectively. Only one abnormal image that belongs to angioectasias class is misclassified as Lymphangiectasia. Also one melanoma image is wrongly classified as other classes and one image from other classes are misclassified as melanoma. Proper classification results from Table VII and VIII are realized using fused results of segmentation and classification. In Fig. 17, binary maps produced from intersection of segmentation and classification maps for a sample bleeding image are illustrated. Binary segmentation and classification maps corresponding to each abnormality-specific layer are presented in top and bottom row of Fig 17, respectively. In the middle row of Fig. 17, intersection of segmentation and classification maps indicates similarity between these two maps in the corresponding abnormality. It can be observed that the intersection results in the bleeding part are similar to both the corresponding segmentation and classification map. In practical experiments, employing only classification map may not lead to appropriate results. Using equation (3), the abnormality correspond to the most similar intersection map is regarded as detected abnormality class. detection are compared with related works respectively. In Table IX, Table X, relative abnormality detection methods have used various features as well as post-processing procedures. However, the proposed method does not use any features and only biggest connected component is used as post-processing procedure.

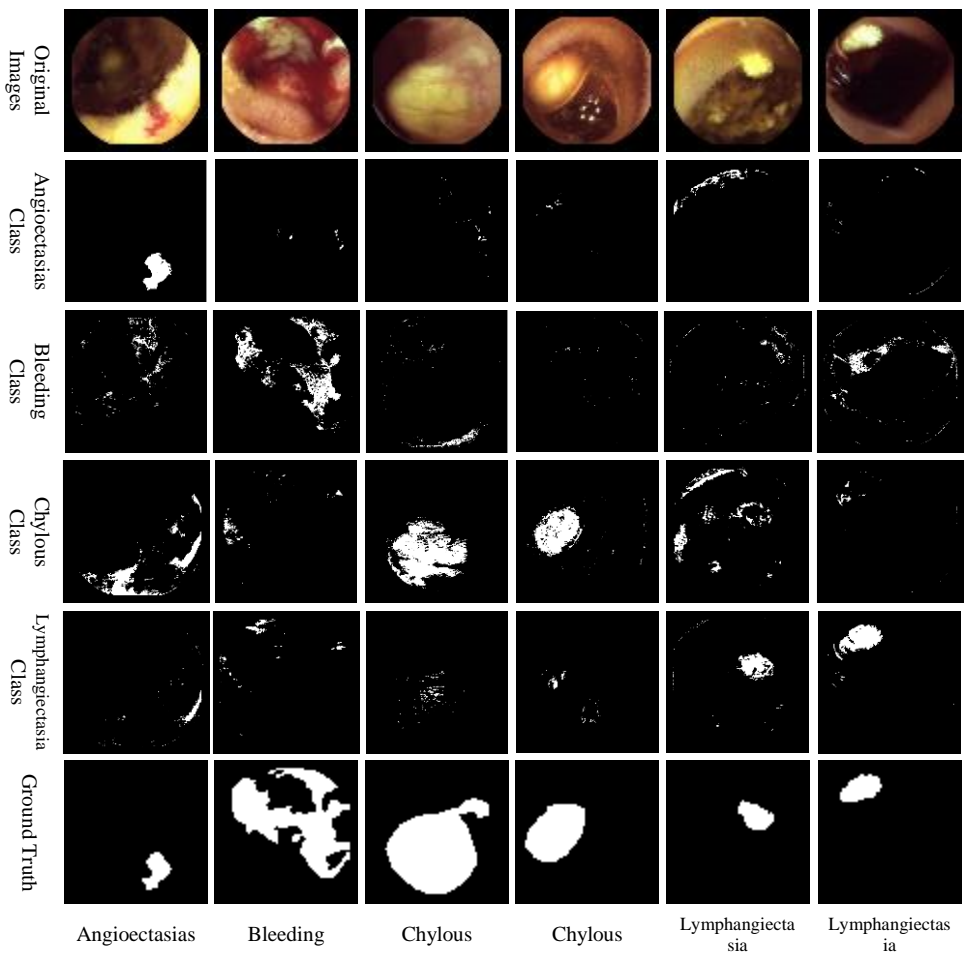

Fig. 13. Results of the classification network with merged primary part for WCE images

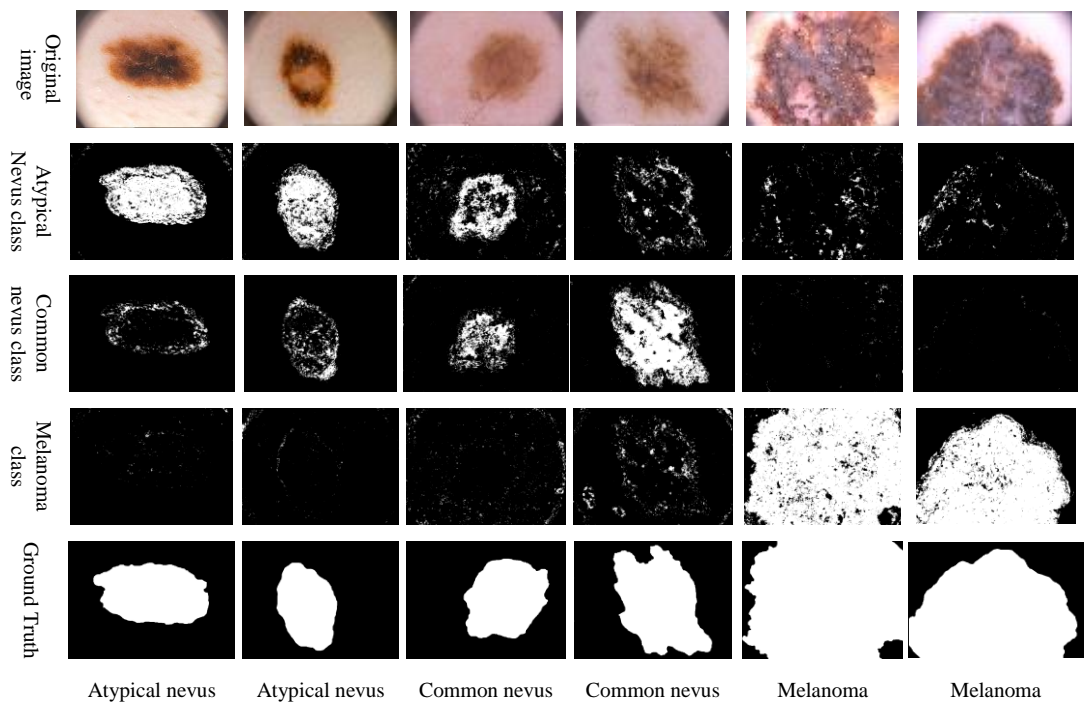

Fig. 14. Results of the classification network with merged primary part for dermoscopic images

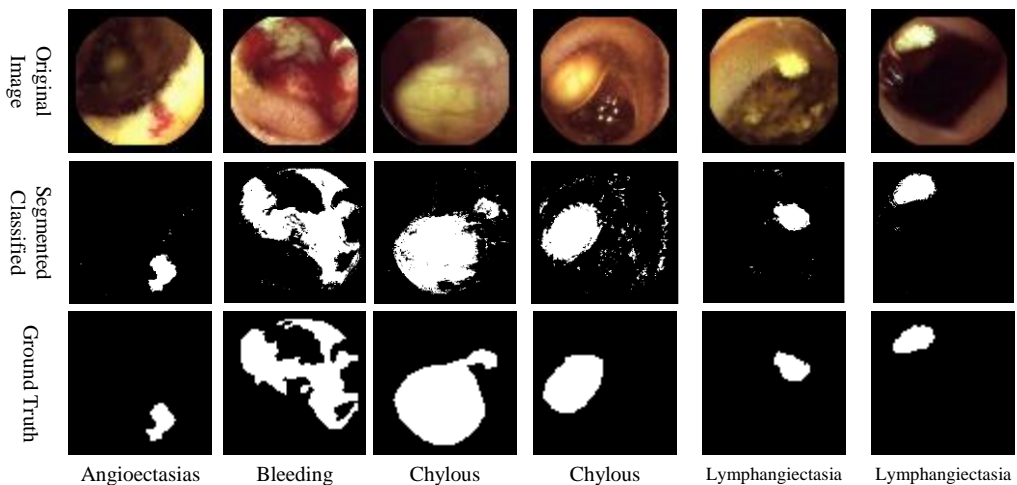

Fig. 15. Final classified-segmented results for WCE images

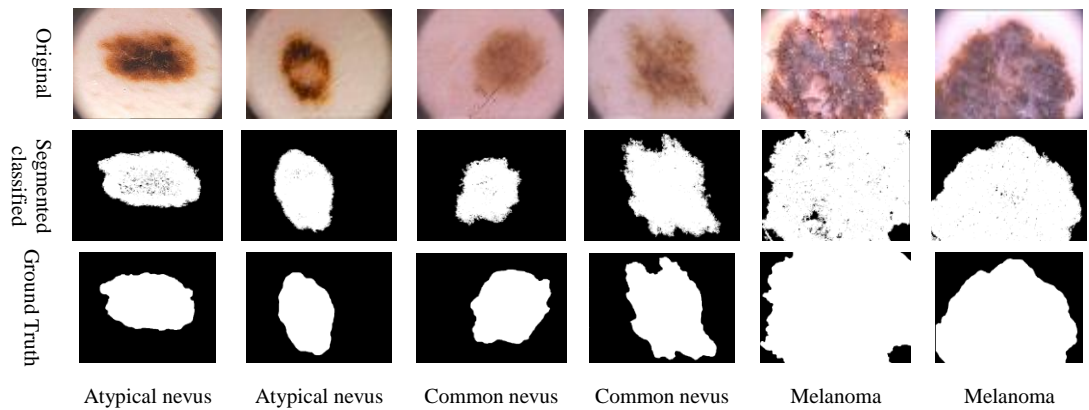

Fig. 16. Final classified-segmented results for dermoscopic images

Table VII.  Number of true and false predictions of WCE abnormality classes

|  | Angioectasias | Bleeding | Chylous | Lymphangiectasia |
|---|---|---|---|---|
| Angioectasias | 26 | 0 | 0 | 1 |
| Bleeding | 0 | 50 | 0 | 0 |
| Chylous | 0 | 0 | 8 | 0 |
| Lymphangiectasia | 0 | 0 | 0 | 9 |

Table VIII.  Number of true and false predictions of skin lesion classes in PH2 dataset

|  | Atypical nevus | Common nevus | Melanoma |
|---|---|---|---|
| Atypical nevus | 72 | 8 | 1 |
| Common nevus | 3 | 77 | 0 |
| Melanoma | 1 | 0 | 39 |

In Table IX and X, quantitative results of the proposed bifurcated network for WCE abnormality and skin lesion In Table IX, Different abnormalities are investigated based on different methods on the WCE images. In [6], maximum posterior (MAP) as well as Markov random field (MRF) are used to segment angioectasias images. In [7], [25] and [26], SVM is utilized for segmentation of bleeding and angioectasias images. In [8] and [9], bleeding WCE images are segmented using SVM. Also in [40], ANN is used for segmentation of bleeding images. In [10] and [27] different types of abnormalities including chylous, bleeding and Lymphangiectasia are classified in image level. The proposed method results are close to the results of the previous methods for all abnormalities. In [10] and [27] image classification is performed and segmentation result is not reported. Our proposed network is able to effectively classify abnormality in

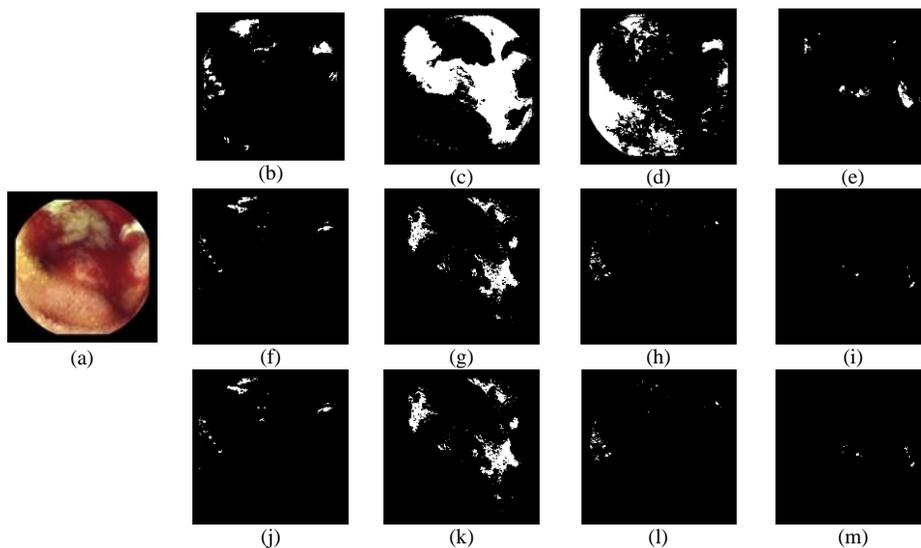

Fig. 17. Binary maps produced from intersection of segmentation and classification maps for a sample input image. (a) input image, (b), (c), (d),(e) binary maps from lymphangiectasia, bleeding, Chylous, and angioectasias networks of the segmentation branch, (f), (g), (h), (i) intersection maps, (j), (k), (l), (m) binary maps from lymphangiectasia, bleeding, Chylous, and angioectasias networks of the classification branch.

 

Table X. Comparative results of the proposed method on dermoscopic images with related works

| | Classification | | | Segmentation | | | |
|---|---|---|---|---|---|---|---|
| | **Sen** | **Spe** | **Acc** | **Sen** | **Spe** | **Acc** | **Dice** |
| [11] , ANN | 0.957 | 100 | 0.967 | -- | -- | -- | -- |
| [12], KNN | 0.960 | 0.830 | -- | -- | -- | -- | -- |
| [13], SVM | 0.977 | 0.967 | 0.975 | -- | -- | -- | -- |
| [15], FCN | -- | -- | -- | -- | -- | -- | 0.938 |
| [16], FCN | -- | -- | -- | 0.948 | 0.939 | 0.942 | 0.906 |
| [17], SVM & RF | 100 | 0.882 | -- | -- | -- | -- | -- |
| [18], region growing | -- | -- | -- | 0.876 | 0.953 | 0.934 | -- |
| [19], SVM | 0.96 | 0.97 | -- | -- | -- | -- | -- |
| [20], SVM | 0.950 | 0.950 | 0.950 | 95.4 | 0.981 | 0.965 | 0.920 |
| [21], Delaunay Triangulation | -- | -- | -- | 0.802 | 0.972 | 0.896 | |
| Monolithic, CNN | -- | -- | -- | 0.724 | 0.978 | 0.891 | 0.820 |
| **Proposed, CNN** | 0.975 | 0.993 | 0.990 | 0.924 | 0.978 | 0.960 | 0.940 |

Table IX. Comparative results of the proposed method on WCE images with related works

| | Angioectasias | Bleeding + Angioectasias | Bleeding | Chylous | Lymphangiectasia |
|---|---|---|---|---|---|
| MAP+MRF [6] | DICE : 0.76 | -- | -- | -- | -- |
| SVM [25] | -- | Sen: 0.91 Spec : 0.96 | -- | -- | -- |
| SVM [7] | -- | Sen: 0.88 Spec : 0.84 | -- | -- | -- |
| SVM [26] | -- | Sen : 0.78 Spec : 0.99 | -- | -- | -- |
| SVM [9] | -- | -- | DICE: 0.84 | -- | -- |
| SVM [8] | -- | -- | DICE: 0.81 | -- | -- |
| ANN [40] | -- | -- | DICE: 0.85 | -- | -- |
| SVM [10] | -- | -- | AUC: 0.83 | AUC : 0.87 | AUC: 0.96 |
| CNN [27] | -- | -- | AUC: 0.64 | AUC : 0.87 | AUC : 0.95 |
| Monolithic CNN | DICE: 0.76 | Sen : 0.46 Spec: 0.99 | DICE: 0.52 | DICE: 0.74 | DICE:0.91 |
| Proposed | DICE: 0.76 | Sen:0.81 Spec: 0.99 | DICE:0.85 | DICE:0.85 | DICE:0.93 |

image level and in Table VII, only one misclassification is observed. For better evaluation of the proposed method, also results of single CNN structure for segmentation is reported in Table IX. It can be observed in Table IX, that the results of the proposed method are better than our single CNN except for angioectasias that similar Dice is obtained. It is important to note that the proposed network can classify and segment different types of abnormalities in a unified structure.

In Table X, [11], [12], [13], [17], [19] and [20], classified skin lesions in malignant and non-malignant classes and in [15], [16], [18], [20] and [21], skin lesion segmentation is investigated. Although in some research studies both segmentation and classification are conducted, only in [20], results of segmentation and classification are reported. In comparison with [20], better classification results and higher Dice score are obtained while achieving. Similar accuracy and specificity. In Table X, Dice score of 0.82 is observed for segmentation in single CNN structure while the proposed bifurcated network has the Dice score of 0.94. In comparison with other related works in Table X, both segmentation and classification tasks are performed in our proposed framework and superior/comparable results are obtained.

## V. CONCLUSION

In this paper, a simple bifurcated neural network structure was designed for classification and segmentation of multiple abnormalities for portable medical instruments. The proposed network was equipped with a primary part to extract common features and two secondary parts for classification and segmentation. Secondary parts could be specialized for each abnormality to achieve better results. Since the proposed structure was designed to be used inside the portable device, it was designed to be simple and benefit from the shared parts of operations. Simulation results were conducted for multiple abnormalities in two medical portable applications including WCE and dermoscopy. It was observed that the proposed network was able to segment and classify WCE and dermoscopic images simultaneously. Also, utilizing primary parts of the network prevented redundant operations which makes it suitable for implementation inside the portable medical devices.